\documentclass{bmvc2k}

\usepackage{booktabs}
\usepackage{xspace}

\usepackage{tabularx}
\newcolumntype{H}{>{\setbox0=\hbox\bgroup}c<{\egroup}@{}}

\usepackage{xcolor,colortbl}
\definecolor{clrfirst}{rgb}{0.95, 0.51, 0.30}
\definecolor{clrsecond}{rgb}{0.96, 0.91, 0.15}
\definecolor{clrthird}{rgb}{0.75, 0.23, 0.0.51}

\usepackage[textsize=small, linecolor=magenta, bordercolor=magenta,
            backgroundcolor=magenta, textwidth=5cm]{todonotes}

\usepackage{pgfplots} 
\pgfplotsset{compat=newest} 
\pgfplotsset{plot coordinates/math parser=false}
\newlength\figheight 
\newlength\figwidth
\usetikzlibrary{shadows}

\expandafter\def\expandafter\normalsize\expandafter{%
    \normalsize
    \setlength\abovedisplayskip{5pt}
    \setlength\belowdisplayskip{5pt}
    \setlength\abovedisplayshortskip{5pt}
    \setlength\belowdisplayshortskip{5pt}
}

\usepackage{glossaries}
\makeglossaries
\newacronym{hr}{{HR}}{high resolution}
\newacronym{lr}{{LR}}{low resolution}
\newacronym{rmse}{{RMSE}}{Root Mean Squared Error}
\newacronym{mae}{{MAE}}{Mean Absolute Error}
\newacronym{tv}{{TV}}{Total Variation}
\newacronym{tgv}{{TGV}}{Total Generalized Variation}
\newacronym{mrf}{{MRF}}{Markov Random Field}
\newacronym{fcn}{{FCN}}{fully-convolutional network}
\newacronym{pdn}{{PDN}}{primal-dual network}

\newcommand{\Method}{Deep Primal-Dual Network\xspace}
\newcommand{\method}{deep primal-dual network\xspace}

\newcommand*{\Figure}{Fig.\@\xspace}
\newcommand*{\Tab}{Tab.\@\xspace}
\newcommand*{\tm}{ToFMark\@\xspace}

\newcounter{mycomment} 

\newcommand*{\etal}{\textit{et al.}\@\xspace} 
\newcommand*{\eg}{\textit{e.g.}\@\xspace}     
\newcommand*{\ie}{\textit{i.e.}\@\xspace}     
\newcommand*{\etc}{\textit{etc.}\@\xspace}    

\newcommand{\sota}{state-of-the-art }

\def\ismpl{k}
\def\nsmpl{K}

\def\dpth{d} 
\def\dpthlr{\dpth^{(\mathrm{lr})}} 
\def\dpthmr{\dpth^{(\mathrm{mr})}} 
\def\dpthhr{\dpth^{(\mathrm{hr})}} 
\def\gdnc{g} 

\def\smpl{s}
\def\smpli{{\smpl_\ismpl}}
\def\trgt{t}
\def\trgti{\trgt_\ismpl}

\def\fcnet{\mathsf{fcn}}
\def\fcneti{\fcnet_\smpli}
\def\fcnetd{\fcnet^{(d)}}
\def\fcnetdi{\fcneti^{(d)}}
\def\fcnetA{\fcnet^{(A)}}
\def\fcnetAi{\fcneti^{(A)}}
\def\pdnet{\mathsf{pdn}}

\def\rglrzr{{R}}
\def\dttrm{{D}}

\def\wgts{w}

\def\sclfctr{\rho}

\def\RealN{\mathbb{R}}

\def\ImgDom{\Omega}

\def\st{\text{s.t.}}
\DeclareMathOperator*{\argmin}{arg\,min}

\DeclareMathOperator{\R}{\mathbb{R}}
\DeclareMathOperator{\nbh}{\mathcal{N}}

\newcommand*\diff{\mathop{}\!\mathrm{d}}

\usepackage{amsmath, amssymb}
\usepackage{graphicx}
\usepackage{epstopdf}

\title{A \Method \\for Guided Depth Super-Resolution}

\addauthor{Gernot Riegler}{riegler@icg.tugraz.at}{1}
\addauthor{David Ferstl}{ferstl@icg.tugraz.at}{1}
\addauthor{Matthias R\"uther}{ruether@icg.tugraz.at}{1}
\addauthor{Horst Bischof}{bischof@icg.tugraz.at}{1}

\addinstitution{
 Institute for Computer Graphics and Vision\\
 Graz University of Technology\\
 Austria
}

\runninghead{Riegler, Ferstl, R\"uther, Bischof}{A \Method}

\begin{document}

\maketitle

\begin{abstract}
In this paper we present a novel method to increase the spatial resolution of depth images.
We combine a deep fully convolutional network with a non-local variational method in a \textit{deep primal-dual network}. 
The joint network computes a noise-free, high-resolution estimate from a noisy, low-resolution input depth map.
Additionally, a high-resolution intensity image is used to guide the reconstruction in the network.
By unrolling the optimization steps of a first-order primal-dual algorithm and formulating it as a network, we can train our joint method end-to-end.
This not only enables us to learn the weights of the fully convolutional network, but also to optimize all parameters of the variational method and its optimization procedure.
The training of such a deep network requires a large dataset for supervision.
Therefore, we generate high-quality depth maps and corresponding color images with a physically based renderer.
In an exhaustive evaluation we show that our method outperforms the \sota on multiple benchmarks.
\end{abstract}

\section{Introduction}
In the last decade, a large range of affordable depth sensors became available on the mass market. 
This has pushed research to develop a variety of different applications based on these sensors.
Especially active sensors based on structured light, or Time-of-Flight (ToF) measurements, enabled novel computer vision applications such as robot navigation~\cite{almansa12}, human pose estimation~\cite{girshick11,shotton11}, and hand pose estimation~\cite{tang14,tang13}.
Despite their success, these sensors suffer from a low spatial resolution and a high acquisition noise due to the 
physical limitations of the measurement principles.
Even very recent ToF sensors have a spatial resolution of only $120\! \times\! 160$ pixels~\cite{pmdpico}.
Therefore, more and more approaches are proposed to improve the resolution and to suppress the noise of these depth cameras. 
Usually these depth cameras are equipped with an additional intensity camera of higher resolution.
Hence, a very common practice~\cite{diebel05,park11,yang07} is to utilize the \gls*{hr} intensity image as guidance.
These approaches build upon the observation that depth discontinuities often occur at high intensity variations and that homogeneous areas in intensity images are also more likely to represent homogeneous areas in depth.

While the classical single image super-resolution for color images is dominated by machine learning approaches, \eg \cite{kim15a,riegler15b,timofte14}, the field of depth super-resolution still mainly relies on Markov Random Field formulations~\cite{diebel05}, adaptive filters~\cite{yang07}, or variational methods~\cite{ferstl13}.
This is due to the lack of high quality training data in larger quantities, which is essential for large-scale machine learning methods.
While it is quite easy to get a huge database of color image examples, \eg from the web, there exists no equivalent source for depth data.
One workaround~\cite{fanello14,kwon15} is to densely reconstruct a 3D scene with KinectFusion~\cite{izadi11} and facilitate these reconstructions as ground-truth.
However, this also introduces artifacts in the training data, such as smoothed edges and the loss of fine details.
Further, the scene preparation for reconstruction and the recording process itself are very time-consuming. 

\begin{figure}[tb]
  \center
  \includegraphics[width=0.98\textwidth]{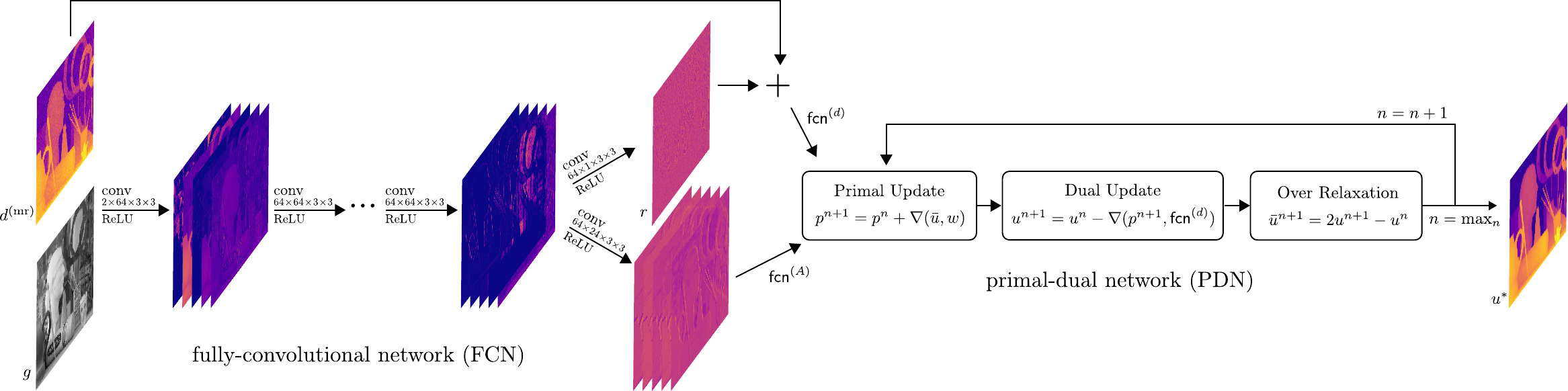}
  \vspace{-10pt}
  \caption{
    Our \textit{\method} consists of two networks. 
    A fully-convolutional network that computes a first \gls*{hr} estimate and weighting coefficients.
    Both outputs are then used in our primal-dual network, where we unroll the optimization steps of a non-local variational method that incorporates prior knowledge about the data modalities.
  }
  \label{fig:architecture}
  \vspace{-14pt}
\end{figure}

In this work we present a novel method based on machine learning for guided depth super-resolution, which combines the advantages of deep convolutional networks and variational methods. 
The training of this novel combination is enabled by creating a large corpus of high quality training data, which are automatically generated by rendering depth maps and corresponding color images from randomly placed and textured 3D objects in a virtual scene.
This data is used to train our \textit{\method} that maps \gls*{lr} and noisy depth maps to accurate \gls*{hr} estimates. 
The first part of the network consists of a series of fully convolutional layers to produce a guidance and rough 
super-resolved depth. This guidance and depth is used in a novel non-local variational model to optimize the final 
result. By unrolling the computation steps of a primal-dual algorithm~\cite{chambolle11} we formulate the variational 
optimization as a \textit{primal-dual network}, where each numerical operation in this algorithm is defined as a layer 
in the network.
In this way, our \textit{\method} enables a joint optimization of all convolutional filter weights, the trade-off parameter of the variational cost function, and all hyper-parameters of the primal-dual algorithm.

The contribution of our work is three-fold and can be summarized as follows:
(i) We extend our work of \cite{riegler16} by combining a deep fully convolutional network with a non-local primal-dual network that is trained end-to-end, shown in Section~\ref{sec:method}.
Hence, we map a noisy, \gls*{lr} depth map along with a \gls*{hr} guidance image to an accurate \gls*{hr} estimate.
(ii) We propose a framework based on the physically based Mitsuba renderer~\cite{wenzel10} to automatically generate 
high-quality depth maps with corresponding color images in large quantities which are used to train our model, shown in 
Section~\ref{sec:training_data}.
(iii) The evaluations presented in Section~\ref{sec:evaluation} demonstrate the effectiveness of our method by 
outperforming \sota results on a set of standard synthetic and real-world benchmarks.

\section{Related Work} 
Single-image super-resolution, \ie enhancing the spatial resolution of an image, is a fundamental problem in low-level computer vision.
It is inherently ill-posed, as several different \gls*{hr} images can map to the very same \gls*{lr} image.
The field can be mainly divided into methods where an edge-preserving smoothness term is utilized~\cite{unger10}, co-occurrences of patches within the same image are exploited~\cite{glasner09}, or, currently most successful, a mapping from \gls*{lr} to \gls*{hr} image patches is learned~\cite{zeyde10,timofte14,schulter15,dong14}.

Although the approaches for single-image super-resolution are quite general, models to increase the spatial resolution of depth maps differ.
First, the modalities of depth maps are different than those in color images.
While color images are characterized by high frequent textures and shading effects, depth data contains more noise, and 
consists of piece-wise affine regions and sharp edges at depth discontinuities.
Second, training data for color images can be easily obtained, explaining the recent success of learning based approaches for single image super-resolution.
Hence, several specific models have been proposed for depth super-resolution.
In the seminal work of Diebel \& Thrun~\cite{diebel05} the super-resolution is formulated as a \gls*{mrf} optimization, where the smoothness prior is weighted by the gradient magnitude of a guidance image.
Park~\etal~\cite{park11} extend this \gls*{mrf} functional by incorporating a non-local means term to better preserve local structures of noisy data.
In~\cite{yang07} Yang~\etal propose an approach that builds upon the assumptions that surfaces are piecewise smooth and pixels with similar color have a similar depth. 
From this they derive a bilateral filter that is iteratively applied to the input depth map.
Similarly, Chan~\etal~\cite{chan08} present a modified bilateral filter to reduce artifacts in areas where a standard 
bilateral upsampling would cause a texture copy.
A variational approach for guided depth super-resolution is proposed by Ferstl~\etal~\cite{ferstl13}.
They formulate the energy functional with an anisotropic Total Generalized Variation prior, which is weighted by the gradients in the guidance intensity image.
Yang~\etal~\cite{yang14} formulate the depth upsampling as a minimization of an adaptive color-guided auto-regressive model.
One of the few learning based approaches is proposed by Kwon~\etal~\cite{kwon15}.
They apply a multi-scale sparse coding approach to iteratively refine the \gls*{lr} depth map, and the \gls*{hr} data for training is acquired with KinectFusion~\cite{izadi11}.

While the approaches discussed above all utilize a \gls*{hr} guidance image, there exist also a few approaches that estimate the \gls*{hr} depth map without guidance. 
Aodha~\etal~\cite{aodha12} and Horn\'{a}\v{c}ek~\etal~\cite{hornacek13} both utilize a MRF to fit \gls*{hr} candidate patches and differ in the search strategy to find similar patches.
Aodha~\etal exploit an external database of a few synthetic depth maps and Horn\'{a}\v{c}ek~\etal search the 3D patches within the same depth map.
A variational depth super-resolution model is presented by Ferstl~\etal~\cite{ferstl15}, where they use sparse coding to estimate the depth discontinuities in the \gls*{hr} depth map.

The integration of energy minimization models, like MRFs or variational methods, into deep networks recently gains a lot of interest.
Chen~\etal~\cite{chen15} show how to integrate a MRF on top of a deep network and train it with back-propagation.
Similarly, Zheng~\etal~\cite{zheng15} unroll the computation steps of the mean field approximation~\cite{kraehenbuehl12} to optimize MRFs on top of a network for semantic segmentation.
They show that the individual computation steps can be realised by operations in a convolutional network.
One of the first integrations of variational models into convolutional networks is proposed by Ranftl \& Pock~\cite{ranftl14} for foreground-background segmentation via implicit differentiation of the energy functional.
Riegler~\etal~\cite{riegler15} use this formulation for guided depth denoising and super-resolution.

Our method is in the spirit of machine learning based approaches, especially it is related to the very deep network of~\cite{kim15a}.
To create a feasible amount of training data, we render high quality depth maps and color images of randomly generated scenes in large quantities.
Like~\cite{riegler15}, we combine the deep network with a variational approach, with the crucial difference that we do not implicitly differentiate the variational method, which drastically limits the choice of energy functionals, but unroll the steps of a fast optimization algorithm, as shown in~\cite{zheng15} for MRFs.
Finally, this work presents several improvments of our method presented in~\cite{riegler16}.
We demonstrate, how to incoorporate an additional guidance image in our method and show that this is crucial for higher upsampling factors.
Gathering training data for this scenario becomes also more difficult.
We solve this problem by using a physically based renderer that produces hiqh quality depth maps along wtih textured color images.
Further, we evaluate different energy functionals for our method and show that a non-local Huber regularation term yields the best trade-off between accuracy and computational requirements for this task.

\section{A \Method} \label{sec:method}

The proposed method consists of two main parts that are jointly trained end-to-end.
The first one is a \gls*{fcn}, which computes a \gls*{hr} estimate of the input depth map and input dependent weighting terms that are utilized in the subsequent \gls*{pdn}.
In the \gls*{pdn} we unroll the optimization procedure of a non-local variational model, namely of the first-order primal-dual algorithm~\cite{chambolle11}.
The unrolling of the optimization steps enables us the integration of the variational model on top of the \gls*{fcn} and a joint training of both networks.
A visual representation of our method is depicted in \Figure~\ref{fig:architecture}.

Let $\dpthlr$ be the \gls*{lr} input depth map with $\dpthlr \in \RealN^{\sclfctr^{-1} M \times \rho^{-1} N}$, where $\sclfctr$ is the scale factor.
The only pre-processing step of our method is an upsampling of $\dpthlr$ via bilinear interpolation to the target resolution.
This yields the mid-resolution input depth map $\dpthmr \in \RealN^{M \times N}$. 
As an additional input we have an intensity image $\gdnc$ as guidance that is given in the target resolution, $\gdnc \in \RealN^{M \times N}$.
For brevity we will condense the mid-resolution depth map and the guidance image to an input sample denoted as $\smpl = (\dpthmr, \gdnc) \in \RealN^{2 \times M \times N}$.
To train our method, we require a dataset $\{(\smpli, \trgti)\}_{\ismpl=1}^\nsmpl$ of $\nsmpl$ input samples $\smpli$ and corresponding \gls*{hr} depth maps as targets $\trgti = \dpthhr_\ismpl \in \RealN^{M \times N}$.
The goal of the model training is to find the optimal parameters $\wgts^* = (\wgts^*_\fcnet, \wgts^*_\pdnet)$ of our model $f = \pdnet(\fcnet(\smpl; \wgts_\fcnet); \wgts_\pdnet)$ that minimize a loss function $L$ over all $K$ training samples:
\begin{align}
  \wgts^* = \argmin_\wgts \sum_{\ismpl=1}^\nsmpl L(f(\smpli, \wgts), \trgti) \,.
\end{align}

\subsection{Fully Convolution Network}
Inspired by~\cite{dong14,kim15a} we use a deep convolutional network to compute an initial high-resolution depth estimate given a noisy, \gls*{lr} input depth map $\dpthlr$ together with a corresponding guidance image $g$.
The network consists of $10$ convolutional layers and rectified linear units (ReLU)~\cite{nair10} as activation functions.
For the convolutional layers we use $3 \times 3$ filter kernels for the benefits discussed in~\cite{simonyan15} and in each convolutional layer we employ $64$ feature maps, which results in a receptive field of $21 \times 21$ pixels of our \gls*{fcn}.

An important aspect of this network is that it does not directly compute the high-resolution depth map $\dpthhr$, but the residual $r = \dpthhr - \dpthmr$ to the mid-resolution input $\dpthmr$ as shown in \Figure~\ref{fig:architecture}.
After addition of the mid-resolution depth to the residual, the network's high-resolution estimate is given by $\fcnetd(\smpli, \wgts_\fcnet) = \dpthmr_\ismpl + r_\ismpl$.
For an uncluttered notation we will denote $\fcnetd(\smpli, \wgts_\fcnet)$ simply as $\fcnetdi$.
The calculation of a residual is especially beneficial for convolutional networks, since it omits the need for the intermediate layers to carry the input information through the whole network, as shown in~\cite{kim15a,schulter15,timofte14}.
This is also related to the recently proposed residual networks for image classification~\cite{he15}.

Additional to the residual output, our network computes weighting coefficients $\fcnetAi$ for the subsequent \gls*{pdn}.
In short, they represent information about depth discontinuities in the \gls*{hr} domain, but we will discuss this in more detail in the next Section.

\subsection{Primal-Dual Network}
As we will show in our evaluations, a \gls*{fcn} already delivers quite satisfying \gls*{hr} estimates for smaller scaling factors.
However, depth dependent noise is still apparent in homogeneous regions. 
In this case variational methods are an ideal solution, since they introduce prior knowledge about the data modalities.
In our method we combine both a \gls*{fcn} and a variational method to estimate sharp and noise-free results all over the image. 
This Section gives the insights into the variational model, how we realize this model as a network and how we combine it with 
the \gls*{fcn} into our complete \textit{\method}.

The cost function of a variational method typically consists of a data term $\dttrm$, which penalizes the deviation from the initial solution and a regularization term $\rglrzr$, where we can formulate smoothness 
assumptions. Hence, our variational model is given by
\begin{align}
  u_\ismpl^* = \argmin_u  \lambda \dttrm(u, \fcnetdi) + \rglrzr(u, \fcnetAi) \,,
  \label{equ:var_model}
\end{align}
where $\dttrm$ and $\rglrzr$ are parametrized by the outputs of the \gls*{fcn}, $\lambda \in \R^+$ steers the weighting between the two terms and $u^*$ is the minimizer of the cost function. 
The data term in our model penalizes the deviations from the \gls*{fcn} depth output and is defined as
\begin{align}
  \dttrm(u, \fcnetdi) =  \frac{1}{2}\int_\ImgDom (u(x) - \fcnetdi(x))^2 \diff x \,.
  \label{equ:var_data}
\end{align}

Most regularization terms are based on first order smoothness assumptions, \eg the \gls*{tv} semi norm, $\rglrzr(u) = \int_\ImgDom \|\nabla u \|_1 \diff x$. 
Although the \gls*{tv} model is able to estimate sharp object discontinuities in the depth map, it has two major disadvantages:
(i) the $\ell_1$ norm favors piecewise constant solutions resulting in piecewise fronto-parallel depth reconstructions. 
(ii) the $\nabla$-operator is not suitable to preserve small scale structures because it only penalizes the forward differences to its direct neighbors.

In our work we model the regularization as \gls*{tv} in a ``larger'' (non-local) neighborhood $\nbh$ and further choose a more robust norm. 
The idea of a non-local regularization~\cite{gilboa09} is to incorporate a low level segmentation process into the variational model. 
This non-local regularization is defined as
\begin{align}
  \rglrzr (u) = \int_\ImgDom \int_{\nbh(x)} w(x,y) | u(x) - u(y) |_\varepsilon  \diff x \diff y, \  \text{where} \ 
  |x|_\varepsilon = [x \le \varepsilon] \tfrac{|x|^2}{2 \varepsilon} + [x > \varepsilon] \left(|x| - \tfrac{\varepsilon}{2}\right),
  \label{equ:var_reg}
\end{align}
where the operator $|\cdot|_\varepsilon$ denotes the Huber norm \cite{huber73}. 
The parameter $\varepsilon \in \R^+$ defines the threshold between the quadratic $\ell_2$ and linear $\ell_1$ norm. 
In contrast to the \gls*{tv} this allows smooth depth reconstruction while preserving sharp edges.
We further call it non-local Huber (NLH) regularization.
One crucial part of this non-local regularization is the weighting factor $w(x,y) \in \R^{|\nbh| \times \Omega}$, which sets the penalty influence of every pixel $y \in \nbh(x)$ to the center $x$. 
The support weight $w(x,y)$ combines the value-similarities and the spatial distances 
\begin{align}
  w(x,y) = \exp{ \left( -\tfrac{\Delta_d}{\sigma_d} - \tfrac{\Delta_a}{\sigma_v}\right) }, 
\end{align}
where $\Delta_d$ denotes the Euclidean proximity $\|x-y\|_2$, which means with increasing distance to $x$ the influence of the penalty decreases. 
$\Delta_a$ denotes the Euclidean affinity for example to a given guidance image, \ie with increasing homogeneity in the guidance also the regularization increases. 
The scalars $\sigma_v, \sigma_d \in \R^+$ define the influence of each term. 

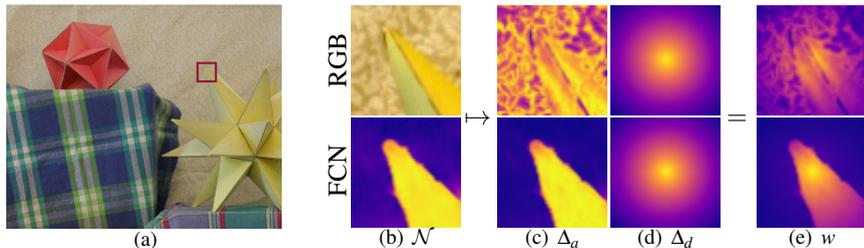
\begin{figure}[tb]
\centering
\newlength{\smallwidth}
\setlength{\smallwidth}{0.12\textwidth}
\setlength{\figheight}{41pt}
\subfigure[]{
  \minipage{0.30\textwidth}
  \includegraphics[trim=350 300 50 0, clip, height=84pt]{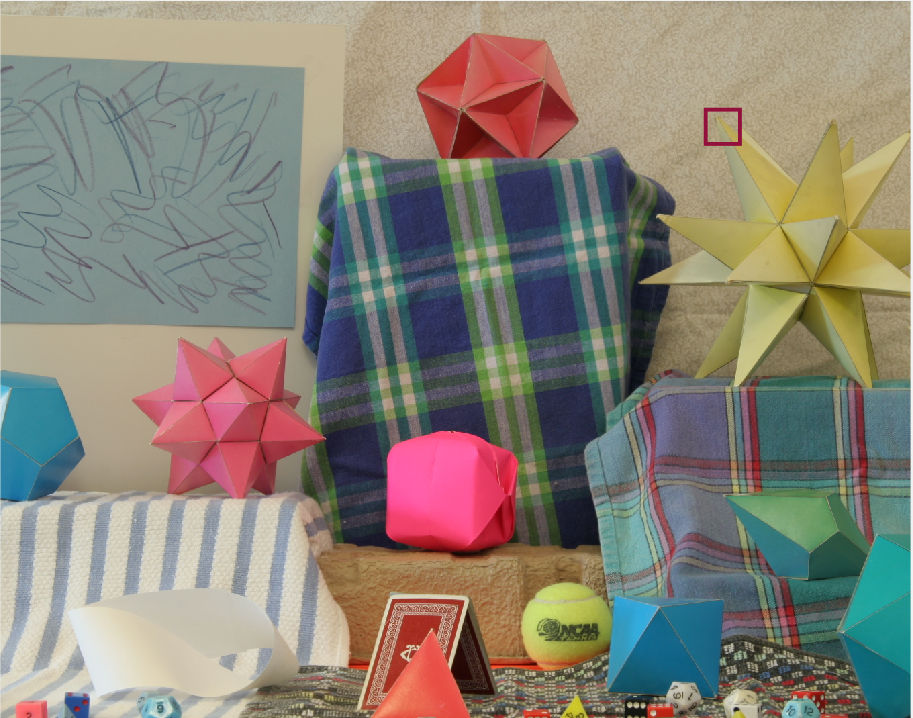}
  \endminipage
}\hspace*{8pt}
\minipage{0.14\smallwidth}
  \begin{picture}(10,10)
    \put(0,16){\rotatebox{90}{RGB}}
    \put(0,-25){\rotatebox{90}{\gls*{fcn}}}
  \end{picture}
\endminipage
\subfigure[$\nbh$]{ 
  \minipage{\smallwidth}
    \includegraphics[height=\figheight]{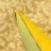}\\
    \includegraphics[height=\figheight]{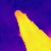}
  \endminipage
}
\minipage{0.13\smallwidth}
  \begin{picture}(10,10)
    \put(-4,2){$\mapsto$}
  \end{picture}
\endminipage
\subfigure[$\Delta_a$]{
  \minipage{\smallwidth}
    \includegraphics[height=\figheight]{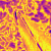}\\
    \includegraphics[height=\figheight]{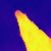}
  \endminipage
} \hspace*{-9pt}
\subfigure[$\Delta_d$]{
  \minipage{\smallwidth}
    \includegraphics[height=\figheight]{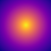}\\
    \includegraphics[height=\figheight]{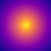}
  \endminipage
}
\minipage{0.13\smallwidth}
  \begin{picture}(10,10)
    \put(-3,2){$=$}
  \end{picture}
\endminipage
\subfigure[$w$]{ 
  \minipage{\smallwidth}
    \includegraphics[height=\figheight]{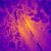}\\
    \includegraphics[height=\figheight]{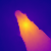}
  \endminipage
}
\vspace*{-5pt}
  \caption{
    NL-support weights. 
    In (a) the input image is shown. 
    (b-e) depict the NL weight calculation. The 1\textsuperscript{st} row shows the traditional calculation from the intensity image and the 2\textsuperscript{nd} row shows the calculation from the trained \gls*{fcn} output. 
    In detail (b) depicts the image neighborhood $\nbh$ from which $w$ is estimated. 
    (c) shows the corresponding affinity part $\Delta_a$ and (d) the proximity part $\Delta_d$. In (e) the final NL-weight matrix $w$ is shown and the advantage of the $\fcnet$ output is clearly visible.
  }
  \label{fig:nl_weights}
\end{figure}

In traditional non-local methods the affinity $\Delta_a$ is given by an intensity image $g$ and results in $\Delta_a = \|g(x) - g(y)\|_2$. 
It has been shown that this is beneficial since high gradients in the intensity image and high depth disparities are likely to co-occur. 
The main disadvantage of this approach is that textured surfaces violate this assumption which subsequently leads to erroneous results. 
Obviously, the optimal guidance would be the high resolution depth image $\dpthhr$. 
Therefore, we use our \gls*{fcn} to directly train for the optimal support weights $w$. Further, since the proximity $\Delta_d$ is constant we only have to train for the affinity term which is defined by the \gls*{fcn} output $\Delta_a = \fcnetA$. 
In \Figure \ref{fig:nl_weights} the difference between the non-local weight from an intensity image and from our learned \gls*{fcn} guidance is shown.  

To optimize the variational model~\eqref{equ:var_model} we use the primal-dual scheme as proposed in~\cite{chambolle11}. 
After discretization of the continuous image space on a Cartesian grid $\Omega \mapsto \R^{M\times N}$ the derived convex-concave saddle-point problem with dual variable $p$ is given by 
\begin{align}
  \min_{u \in \R^{M\times N}} \max_{p \in \mathcal{P}} \left\lbrace \sum_{x \in \R^{M\times N}} \sum_{y \in \nbh(x)} \left(u(x) - u(y)\right) p(x,y) + \frac{\lambda}{2} \| u - \fcnetd \|_2^2 - \frac{\varepsilon}{2} \| p \|_2^2 \right\rbrace  \\
  \st \; p \in \mathcal{P} = \left\lbrace p\colon \R^{M\times N} \mapsto \R \big| |p(x,y)| \leq w(x,y), \; \forall x \in \R^{M\times N}, y \in \nbh(x) \right\rbrace \,.
\end{align}
The iterations of the primal-dual scheme are then given by
\begin{align}
  \begin{cases}
    p^{n+1}(x,y) &= \max \left( -w(x,y), \min \left( w(x,y), \dfrac{p^n(x,y) + \sigma_p (\bar{u}(x) - \bar{u}^n(y))}{1 + \sigma_p \varepsilon} \right) \right)\\
    u^{n+1}(x) &= \dfrac{u^n(x) - \tau_u \left( \sum_{y \in \nbh(x)} p^{n+1}(x,y) - p^{n+1}(y,x) + \lambda \fcnetd(x)\right)}{1 + \tau_u \lambda}\\
    \bar{u}^{n+1}(x) &= 2 u^{n+1}(x) - u^{n}(x)
  \end{cases}.
  \label{eq:pd_opt_scheme}
\end{align}

In traditional primal-dual optimization \eqref{eq:pd_opt_scheme} is solved iteratively, the time-steps $\tau_u, \sigma_p$ are set to be Lipschitz continuous, and the parameters $\lambda, \sigma_d, \sigma_v, \varepsilon$ in the model are searched empirically. 
In contrast, we formulate the whole variational primal-dual optimization as our \glsdesc*{pdn}. 
Hence, each operation in the optimization (addition, multiplication, division, \etc) is defined as a network layer and a fixed number of iterations is \textit{unrolled}, similar as in recurrent neural networks. 
Compared to standard primal-dual optimization our \gls*{pdn} has the advantages that it not only optimizes each parameter of the model in each iteration separately, but also trains separate time-steps for each iteration which are not tied to conservative Lipschitz boundaries.

\subsection{Training}
In general, we train our method by stochastic gradient descent with an additional momentum term.
It is possible to randomly initialize the weights of our model and then train it from scratch.
However, in practice we observed faster convergence and increased accuracy, if we pre-train the \gls*{fcn} in advance.
Therefore, we train the \gls*{fcn} for $25$ epochs with a constant learning rate of $10^{-3}$ and momentum term set to $0.9$ minimizing 
\begin{align}
  \sum_{\ismpl=1}^\nsmpl \sum_{x \in \R^{M\times N}} ||\fcnetdi(x) - \trgti(x)||_2^2 + \sum_{y \in \nbh(x)} ||\fcnetAi(x, y) - (\trgti(x) - \trgti(y))||_\varepsilon \,.
\end{align}
After the pre-training step, we plug $20$ iterations of our \gls*{pdn} on top of the \gls*{fcn} and train both networks jointly for $10$ epochs at a learning rate of $10^{-4}$, minimizing the Euclidean loss.
We note that the loss function can easily be changed, to evaluate different metrics than the \gls*{rmse}.
In this joint training the parameters of the \gls*{fcn} adapt to the \gls*{pdn}, \ie the outputs $\fcnetd$ and $\fcnetA$ get optimized to increase the overall accuracy.
Additionally, all parameters of the \gls*{pdn} improve as well.
This includes the trade-off parameter $\lambda$ and all hyper-parameters of the primal-dual algorithm.
Especially, the parameters get tuned for each iteration individually, yielding an optimal convergence for a fixed number of \gls*{pdn} iterations.

\begin{figure}[tb]
  \center
  \setlength{\figwidth}{0.157\textwidth}
  \includegraphics[width=\figwidth]{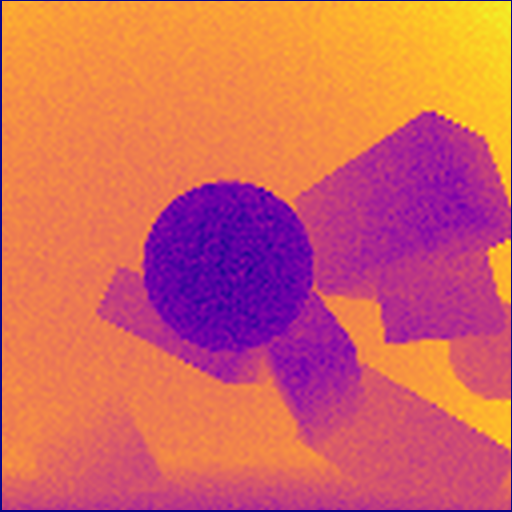}\hspace*{-1pt}
  \includegraphics[width=\figwidth]{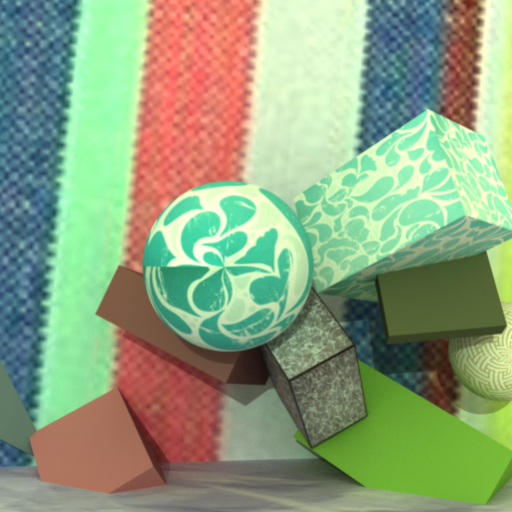}\hspace*{-1pt}
  \includegraphics[width=\figwidth]{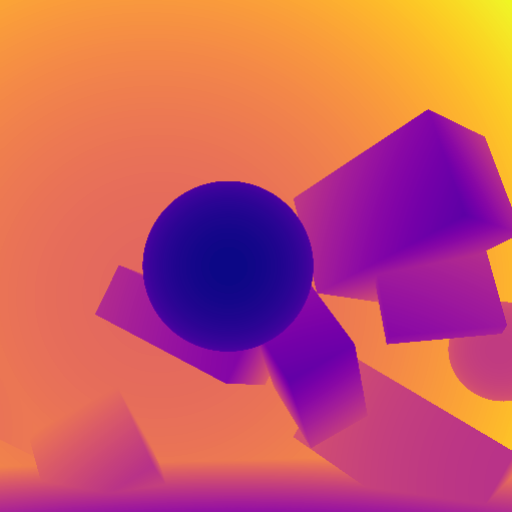}
  \hspace*{8pt}
  \includegraphics[width=\figwidth]{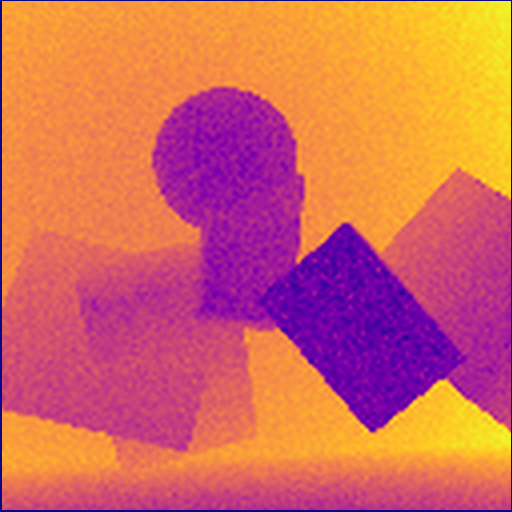}\hspace*{-1pt}
  \includegraphics[width=\figwidth]{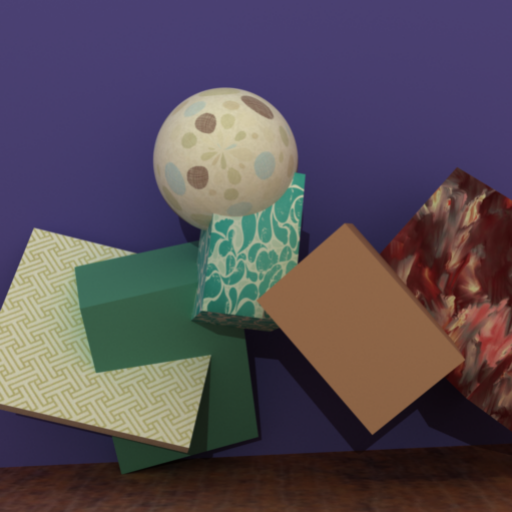}\hspace*{-1pt}
  \includegraphics[width=\figwidth]{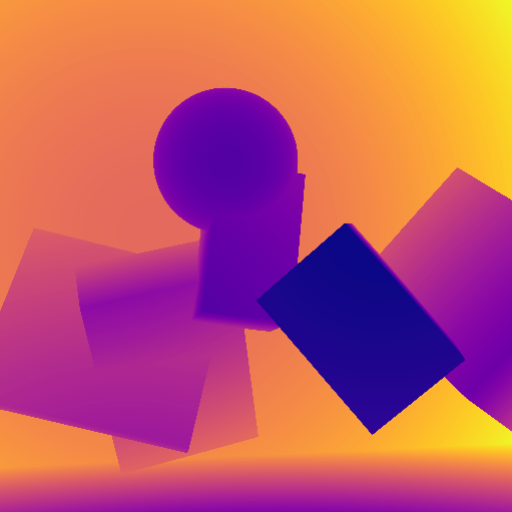}
  \vspace*{-7pt}
  \caption{
    Using a physically based renderer we automatically generate 3D scenes of random objects varying in position, size and texture.
    We also randomly change the lighting directions and intensities. 
    Each sample consists of a noisy \gls*{lr} depth, a \gls*{hr} guidance and a \gls*{hr} target.
  }
  \label{fig:training_data}
  \vspace{-6pt}
\end{figure}

\section{Training Data} \label{sec:training_data}
In this Section we show how we automatically generate our training data. 
Each training sample $(\smpli, \trgti)$ is generated using the open source Mitsuba Render Software~\cite{wenzel10}.
In this physically based renderer a scene is defined by placing objects, light sources and sensors freely in an environment defined by a configuration file. 
Using this file, the renderer generates an intensity- and a depth-map of the scene in definable quality and size. 

In our case, the automatic dataset generation is scripted by randomly placing different objects (cubes, spheres and 
planes) in varying poses and dimensions in the scene. 
Further, the objects are randomly textured using samples from the publicly available \textit{Describable Textures Dataset}~\cite{cimpoi14}.
The light intensity and position is also slightly varied during the data generation. 
We depict two such generated samples in \Figure~\ref{fig:training_data}. 
The output intensity image is used as \gls*{hr} guidance image $\gdnc$, the clean depth output defines the \gls*{hr} target depth $\trgti$, and by downsampling $\trgti$ and adding noise we generate the \gls*{lr} depth input $\dpthlr$.

\section{Evaluation} \label{sec:evaluation}
In the following Section we present a comprehensive evaluation of our \textit{\method}.
First, we demonstrate the influence of different energy functionals and the non-local window size on our \gls*{pdn}.
Then, we compare our method to \sota approaches for guided depth super-resolution on the Middlebury dataset as proposed by Park~\etal~\cite{park11}.
Finally, we present our results on the challenging \tm benchmark~\cite{ferstl13} for real Time-of-Flight data.

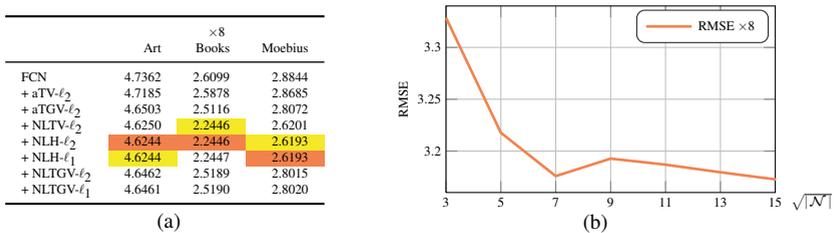
\begin{figure}[tb]
\vspace*{-5pt}
\newcommand{\white}[1]{{\textcolor{white}{#1}}}
  \center
  \subfigure[]{
\minipage{0.35\textwidth}
    \vspace*{3pt}
    {\tiny \begin{tabular}{l r r r } \toprule 
& \multicolumn{3}{c}{$\times 8$} \\  & Art & Books & Moebius \\ \midrule 
\gls*{fcn} & 4.7362 & 2.6099 & 2.8844 \\ 
+ aTV-$\ell_2$ & 4.7185 & 2.5878 & 2.8685 \\ 
+ aTGV-$\ell_2$ & 4.6503 & 2.5116 & 2.8072 \\ 
+ NLTV-$\ell_2$ & 4.6250 & \cellcolor{clrsecond}2.2446 & 2.6201 \\ 
+ NLH-$\ell_2$ & \cellcolor{clrfirst}4.6244 & \cellcolor{clrfirst}2.2446 &\cellcolor{clrsecond} 2.6193 \\ 
+ NLH-$\ell_1$ & \cellcolor{clrsecond}4.6244 & 2.2447 & \cellcolor{clrfirst}2.6193 \\ 
+ NLTGV-$\ell_2$ & 4.6462 & 2.5189 & 2.8015 \\ 
+ NLTGV-$\ell_1$ & 4.6461 & 2.5190 & 2.8020 \\ 
\bottomrule\end{tabular} 
}
    \vspace*{2pt}
  \label{tab:results_diffopt}
\endminipage
}
\hspace{10pt}
  \subfigure[]{
\minipage{0.42\textwidth}
\setlength{\figwidth}{130pt}
\setlength{\figheight}{70pt}
  {\tiny 
%
%
\begin{tikzpicture}

\begin{axis}[%
width=0.951\figwidth,
height=\figheight,
at={(0\figwidth,0\figheight)},
scale only axis,
xmin=3,
xmax=15,
xtick={3,5,7,9,11,13,15,17},
x label style={at={(axis description cs:1.11,0.02)},anchor=north},
xlabel={$\sqrt{|\nbh|}$},
xmajorgrids,
ymin=3.16,
ymax=3.34,
ylabel={RMSE},
ymajorgrids,
axis background/.style={fill=white},
legend style={legend cell align=left,align=left,draw=white!15!black,rounded corners=3pt,
    }
]
\addplot [color=clrfirst,solid,line width=1.0pt]
  table[row sep=crcr]{%
3	3.32868534514779\\
5	3.21765956843586\\
7	3.17582779568674\\
9	3.19270813353978\\
11	3.18687791427015\\
13	3.17952665347244\\
15	3.17267465255878\\
17	3.16668180564006\\
};
\addlegendentry{RMSE $\times 8$};

\end{axis}
\end{tikzpicture}%
  }
\vspace*{-13pt}
\endminipage
  \label{fig:results_ws}
  }
\vspace*{-6pt}
  \caption{
    Influence of the variational model (a) and the non-local neighborhood size $\nbh$ (b) on the \gls*{rmse}.  
    Best results highlighted in \textcolor{clrfirst}{orange} and second best in \textcolor{clrsecond}{yellow}.
  }
  \label{tab:results_diffoptws}
\end{figure}

\begin{table}[tb]
  \center
  {\tiny
    \begin{tabular}{l r r r r r r r r r r r r } \toprule
 & \multicolumn{3}{c}{$\times 2$} & \multicolumn{3}{c}{$\times 4$} & \multicolumn{3}{c}{$\times 8$} & \multicolumn{3}{c}{$\times 16$} \\ \cmidrule(lr){2-4} \cmidrule(lr){5-7} \cmidrule(lr){8-10} \cmidrule(lr){11-13} 
  & Art & Books & Moebius & Art & Books & Moebius & Art & Books & Moebius & Art & Books & Moebius\\ \midrule
                            NN & 6.55 & 6.16 & 6.59 & 7.48 & 6.31 & 6.78 & 9.02 & 6.62 & 7.00 & 11.45 & 7.33 & 7.52 \\
                      Bilinear & 4.58 & 3.95 & 4.20 & 5.62 & 4.31 & 4.56 & 7.14 & 4.71 & 4.87 & 9.72 & 5.38 & 5.43 \\
      Yang \etal \cite{yang07} & 3.01 & 1.87 & 1.92 & 4.02 & 2.38 & 2.42 & 4.99 & 2.88 & 2.98 & 7.85 & 4.27 & 4.40 \\
          He \etal \cite{he10} & 3.55 & 2.37 & 2.48 & 4.41 & 2.74 & 2.83 & 5.72 & 3.42 & 3.57 & 8.49 & 4.53 & 4.58 \\
Diebel \& Thrun \cite{diebel05} & 3.49 & 2.06 & 2.13 & 4.51 & 3.00 & 3.11 & 6.39 & 4.05 & 4.18 & 9.39 & 5.13 & 5.17 \\
      Chan \etal \cite{chan08} & 3.44 & 2.09 & 2.08 & 4.46 & 2.77 & 2.76 & 6.12 & 3.78 & 3.87 & 8.68 & 5.45 & 5.57 \\
      Park \etal \cite{park11} & 3.76 & 1.95 & 1.96 & 4.56 & 2.61 & 2.51 & 5.93 & 3.31 & 3.22 & 9.32 & 4.85 & 4.48 \\
 Ferstl \etal \cite{ferstl13}  & 3.19 & 1.52 & 1.47 & 4.06 & 2.21 & 2.03 & 5.08 & 2.47 & \cellcolor{clrsecond}2.58 & 7.61 & \cellcolor{clrsecond}3.54 & \cellcolor{clrfirst}3.50 \\\midrule
                FCN($\dpthmr$) & \cellcolor{clrsecond}1.83 & 1.10 & 1.26 & 3.03 & 1.73 & 1.99 & 5.39 & 2.65 & 3.08 & 9.31 & 4.34 & 4.40 \\
            FCN+NLH($\dpthmr$) & 2.10 & 1.25 & 1.38 & \cellcolor{clrsecond}2.95 & 1.63 & 1.88 & 5.31 & 2.40 & 2.91 & 9.29 & 4.08 & 4.18 \\
            FCN-PDN($\dpthmr$) & \cellcolor{clrfirst}1.81 & \cellcolor{clrsecond}1.05 & \cellcolor{clrsecond}1.21 & \cellcolor{clrfirst}2.85 & \cellcolor{clrfirst}1.53 & \cellcolor{clrsecond}1.74 & 5.20 & 2.26 & 2.68 & 8.68 & 3.70 & 3.99 \\\midrule
                  FCN($\smpl$) & 1.99 & 1.20 & 1.37 & 3.25 & 1.78 & 1.96 & 4.74 & 2.61 & 2.88 & 7.80 & 4.08 & 4.16 \\
              FCN+NLH($\smpl$) & 2.00 & 1.18 & 1.31 & 3.26 & 1.62 & 1.83 & \cellcolor{clrsecond}4.63 & \cellcolor{clrsecond}2.25 & 2.62 & \cellcolor{clrsecond}7.60 & 3.59 & 3.84 \\
              FCN-PDN($\smpl$) & 1.87 & \cellcolor{clrfirst}1.01 & \cellcolor{clrfirst}1.16 & 3.11 & \cellcolor{clrsecond}1.56 & \cellcolor{clrfirst}1.68 & \cellcolor{clrfirst}4.48 & \cellcolor{clrfirst}2.24 & \cellcolor{clrfirst}2.48 & \cellcolor{clrfirst}7.35 & \cellcolor{clrfirst}3.46 & \cellcolor{clrsecond}3.62 \\
\bottomrule\end{tabular}

  }
  \caption{
    Quantitative results on noisy Middlebury data: 
    We present our results on the disparity maps of the noisy Middlebury dataset~\cite{park11} as \gls*{rmse} of the disparity values.
    Best results highlighted in \textcolor{clrfirst}{orange} and second best in \textcolor{clrsecond}{yellow}.
  }
  \label{tab:results_noisy_middlebury}
  \vspace*{-6pt}
\end{table}

\subsection{Influence of Energy Functional and Non-Local Window Sizes}
In this evaluation we show the influence of the variational model and non-local window size on the accuracy of our model. 
First, we optimize a variety of different variational models without joint training on top of the \gls*{fcn} output. 
The \gls*{rmse} accuracy is shown in \Figure~\ref{tab:results_diffopt} evaluated on the noisy Middlebury data~\cite{park11} ($\times\! 8$) for $20$ iterations. 
We compare two local models, the anisotropic \gls*{tv} and the anisotropic \gls*{tgv} with $\ell_2$ data term~\cite{ferstl13}, and the non-local models with \gls*{tv}, Huber and the recently proposed non-local \gls*{tgv} regularization~\cite{ranftl14b} with $\ell_1$ and $\ell_2$ data term.
The neighborhood size $\nbh$ is set to $7\! \times\! 7$. 
We can observe that all variational models increase the final accuracy. 
While the influence of the data penalization is not very significant, the non-local regularization has a superior performance over the local models. 
Overall, the non-local Huber regularization gives the best results.
Second, we evaluate the influence of the non-local window size on the accuracy in \Figure~\ref{fig:results_ws}. 
The error decreases with a larger neighborhood $\nbh$, but also the computational complexity and memory requirements increase dramatically.
Hence, we use a $7\! \times\! 7$ window size since it provides the best trade-off between accuracy and computational resources.

\vspace*{-4pt}
\subsection{Noisy Middlebury}
\vspace*{-2pt}

\begin{figure}[tb]
\setlength{\figwidth}{0.23\textwidth}
  \centering
  \subfigure[GT and Input]{
    \begin{minipage}{\figwidth} \centering
      \includegraphics[width=\textwidth]{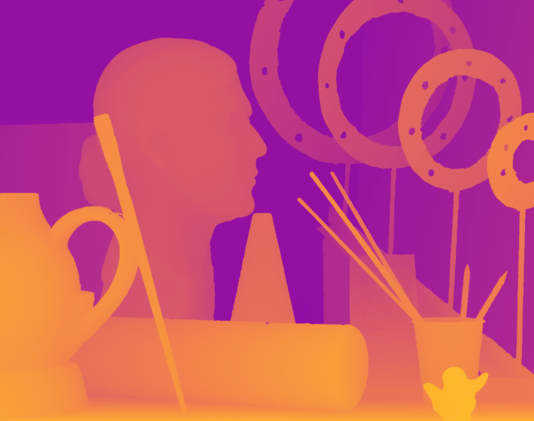}
      \includegraphics[width=\textwidth]{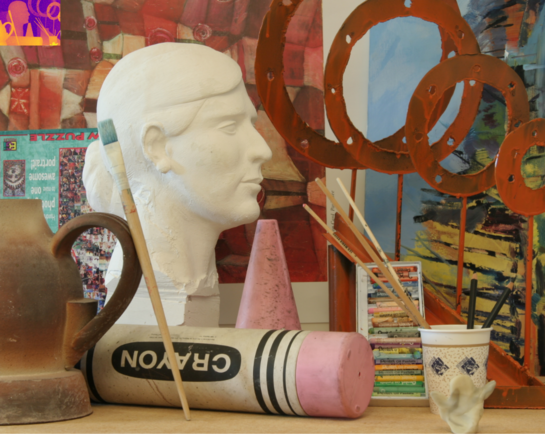}
    \end{minipage} 
  }
  \subfigure[Ferstl~\etal~\cite{ferstl13}]{
    \begin{minipage}{\figwidth} \centering
      \includegraphics[width=\textwidth]{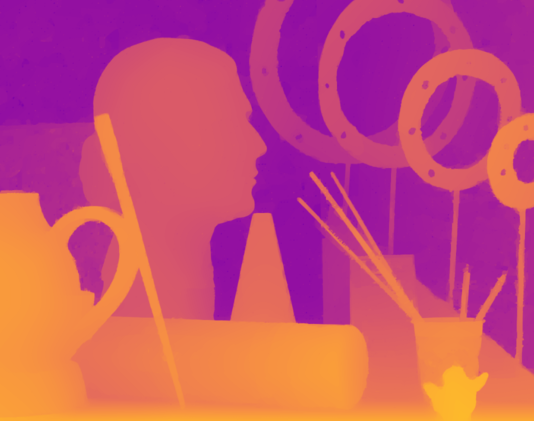}
      \includegraphics[width=\textwidth]{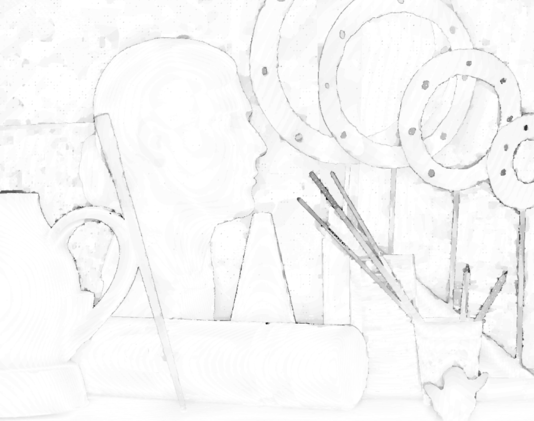}
    \end{minipage} 
  }
  \subfigure[FCN($\smpl$)]{
    \begin{minipage}{\figwidth} \centering
      \includegraphics[width=\textwidth]{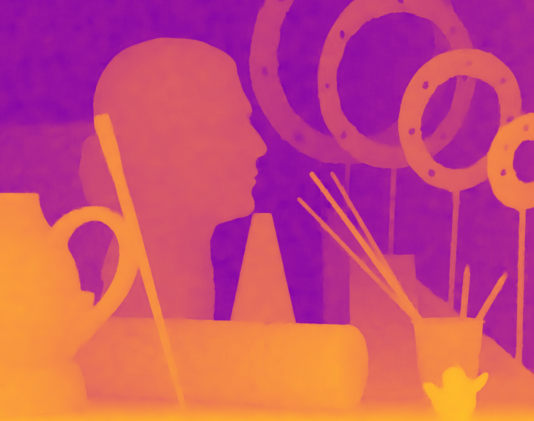}
      \includegraphics[width=\textwidth]{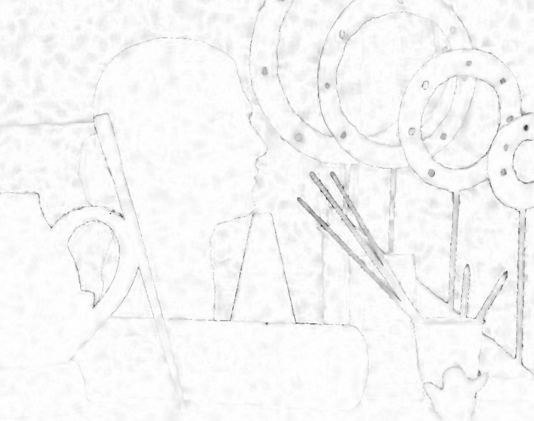}
    \end{minipage} 
  }
  \subfigure[FCN-PDN($\smpl$)]{
    \begin{minipage}{\figwidth} \centering
      \includegraphics[width=\textwidth]{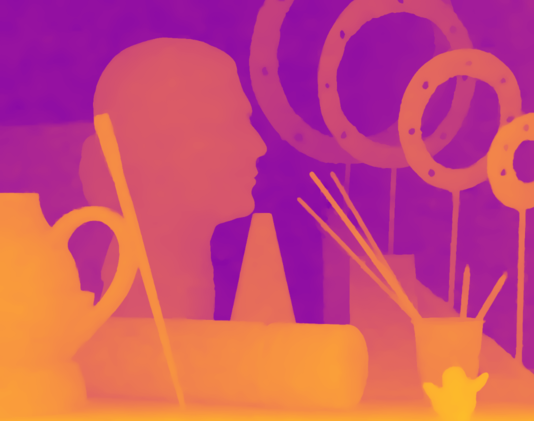}
      \includegraphics[width=\textwidth]{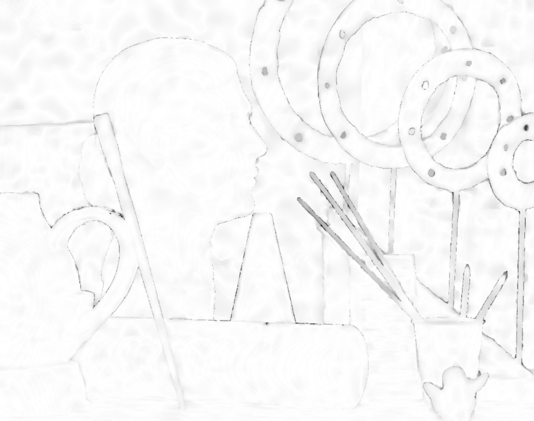}
    \end{minipage} 
  }
\vspace{-4pt}
  \caption{
    Qualitative results for the image \textit{Art} from the noisy Middlebury dataset~\cite{park11} and a scale factor of $\times 8$.
    The first image in (a) shows the ground-truth HR depth and the second image depicts the input sample.
    In (b)-(c) we present in the first row the HR estimates of a \sota method, as well as our results, and in the second row we show the corresponding error maps.
  }
  \label{fig:nmb_qual_res}
  \vspace{-7pt}
\end{figure}

In the following experiment we evaluate our method on the noisy Middlebury dataset as proposed by Park~\etal~\cite{park11}.
According to \cite{park11} we interpret the disparity values as depth.
The disparity maps are corrupted by multiplicative Gaussian noise $\eta(x) = \mathcal{N}(0, 651 \cdot \dpthlr(x)^{-1})$.
The same noise is added to our training data. 
In \Tab~\ref{tab:results_noisy_middlebury} we compare our method to standard interpolation methods and a variety of \sota methods for guided depth super-resolution.
Further, we compare our method once trained solely on the depth maps as input ($\dpthmr$) and once with the additional guidance image as input ($\smpl$).
In this comparison we also show the results of the \gls*{fcn} output only (FCN), the results after applying the variational NLH-$\ell_2$ on top of the \gls*{fcn} (FCN+NLH), and after joint training of our \textit{\method} (FCN-PDN).
For smaller upsampling factors ($\times 2$, $\times 4$) the \gls*{fcn} alone already outperforms all other \sota methods on this dataset, and our complete method after joint training performs best.
At smaller upsampling factors the additional guidance input is not beneficial to the accuracy, but this changes drastically for higher upsampling factors ($\times 8$, $\times 16$).
There, we can observe a significant boost in performance by adding the guidance input.
In those cases the \gls*{pdn} clearly improves the results over the \gls*{fcn} alone.
In \Figure~\ref{fig:nmb_qual_res} we show a example of the qualitative results. We refer to the supplemental material 
for more visualizations.

\subsection{\tm}

\begin{figure}[tb]
\setlength{\figheight}{58pt}
  \center
  \hspace*{-5pt}
  \subfigure[]{
    {\tiny
      \begin{tabular}{l r r r } \toprule
  & Books & Devil & Shark\\ \midrule
                            NN & 30.46 & 27.53 & 38.21 \\
                      Bilinear & 29.11 & 25.34 & 36.34 \\
      Kopf \etal \cite{kopf07} & 27.82 & 24.30 & 34.79 \\
          He \etal \cite{he10} & 27.11 & 23.45 & 33.26 \\
 Ferstl \etal \cite{ferstl13}  & \cellcolor{clrsecond}24.00 & \cellcolor{clrsecond}23.19 & \cellcolor{clrsecond}29.89 \\\midrule
FCN-PDN ($\dpthmr$ \& $\gdnc$) & \cellcolor{clrfirst}23.74 & \cellcolor{clrfirst}20.47 & \cellcolor{clrfirst}28.81 \\
\bottomrule\end{tabular}

    }
  }\hspace*{-7pt}
  \subfigure[{\scriptsize \gls*{fcn}($\smpl$)}]{
    \minipage{0.197\textwidth}
      \vspace*{4pt}
      \includegraphics[height=\figheight]{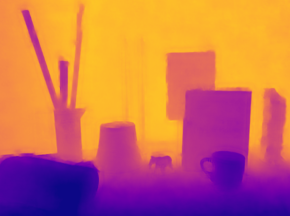}
      \vspace*{-11pt}
    \endminipage
  }
  \subfigure[{\scriptsize FCN-PDN($\smpl$)}]{
    \minipage{0.197\textwidth}
    \vspace*{4pt}
      \includegraphics[height=\figheight]{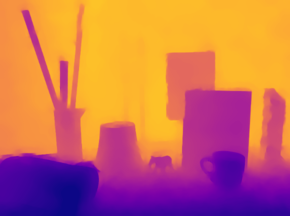}
    \vspace*{-11pt}
    \endminipage
  }
  \subfigure[{\scriptsize $\dpthhr$}]{
    \minipage{0.197\textwidth}
    \vspace*{4pt}
      \includegraphics[height=\figheight]{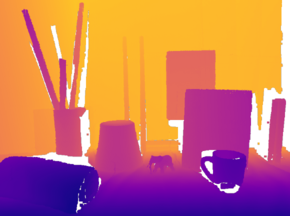}
    \vspace*{-11pt}
    \endminipage
  }
  \vspace*{-5pt}
  \caption{
    Quantitative and qualitative results on the \tm~\cite{ferstl13} benchmark.
    In (a) we present our quantitative results as \gls*{rmse} in $mm$.
    Best results highlighted in \textcolor{clrfirst}{orange} and second best in \textcolor{clrsecond}{yellow}.
    In (b) and (c) we show the results of the \gls*{fcn} and the full model, respectively.
    For comparison, we also show in (d) the ground-truth \gls*{hr} depth.
  }
  \label{tab:results_tofmark}
\end{figure}

In our final evaluation we compare our method on the challenging real-world \tm dataset~\cite{ferstl13}.
The dataset consists of three different scenes. For each scene it provides a noisy, \gls*{lr} ToF image, a \gls*{hr} depth map, generated with a structure light scanner, and a \gls*{hr} intensity image.
The intensity image and the \gls*{hr} depth map are in the same camera coordinate system, however, the \gls*{hr} depth map is given in its own system.
Therefore, the depth pixels are mapped to the \gls*{lr} coordinate system of the intensity image via the provided projection matrix. 
This yields a sparse depth map which we fill with bilinear interpolation to generate the mid-resolution input. 
In the training data we simulated this projection. 
First, the \gls*{hr} training depth maps are mapped in the \gls*{lr} ToF coordinate system via the inverse projection matrix.
Since multiple \gls*{hr} points can map onto the same \gls*{lr} pixel, we compute the mean over the corresponding depth values. 
Second, we apply depth dependent noise on the \gls*{lr} depth from which the mid-resolution input is generated. 
In \Tab~\ref{tab:results_tofmark}a we compare the results of our method with \sota guided depth super-resolution methods, where we can observe a significant improvement in terms of the root mean squared error over previous approaches.
A qualitative result is depicted in \Figure~\ref{tab:results_tofmark}b-d.
Again, we refer to the supplemental material for more visualizations.

\section{Conclusion}
We presented a novel method that combines the advantages of deep fully convolutional networks and variational methods for guided depth super-resolution.
We formulated the non-local variational model as a network which is placed on top of a fully convolutional network by unrolling the optimization steps of a primal-dual algorithm.
In a complete end-to-end training our \textit{\method} is able to learn an efficient parameterization of the model 
including the convolutional filters, and all hyper-parameter and step-sizes of the variational optimization.
We created the necessary training data with a physically based renderer in high quality and large quantities.
In our evaluations we have shown that this novel combination significantly outperforms \sota results on different 
synthetic and real-world benchmarks.

\paragraph*{Acknowledgments} 
This work was supported by \emph{Infineon Technologies Austria AG} and the Austrian Research Promotion Agency (FFG) under the \emph{FIT-IT Bridge} program, project \#838513 (TOFUSION).

\bibliography{index}
\end{document}